\documentclass[letterpaper, 10 pt, conference]{ieeeconf}

\usepackage[utf8]{inputenc}
\usepackage[T1]{fontenc}
\usepackage{acronym}

\usepackage{array}
\usepackage{float}
\usepackage{amsfonts}
\usepackage{amsmath}
\usepackage{amssymb}
\usepackage{mathtools}

\DeclarePairedDelimiterX{\infdivx}[2]{(}{)}{%
	#1\;\delimsize\|\;#2%
}

\DeclarePairedDelimiter{\norm}{\lVert}{\rVert}

\usepackage[english]{babel}

\usepackage{wrapfig}
\usepackage[export]{adjustbox}
\usepackage{microtype}
\usepackage[dvipsnames]{xcolor}
\usepackage{graphicx}
\usepackage{graphics}
\usepackage{svg}
\usepackage{tikz}

\usepackage{multicol}
\usepackage{multirow}
\usepackage[font=footnotesize]{caption}
\usepackage{subcaption}
\usepackage{booktabs}

\usepackage{authblk}
\usepackage{algorithm, algorithmicx}
\usepackage[noend]{algpseudocode}

\usepackage{todonotes}
\usepackage{listings}

\renewcommand{\thetable}{\arabic{table}}

\graphicspath{{/images}}
\newcommand\red[1]{\textcolor{red}{#1}}

\makeatletter
\def\endthebibliography{%
	\def\@noitemerr{\@latex@warning{Empty `thebibliography' environment}}%
	\endlist}
\makeatother

\definecolor{label-running} {RGB}{ 31,119,180}
\definecolor{label-walking} {RGB}{255,127, 14}
\definecolor{label-jumping} {RGB}{ 44,160, 44}
\definecolor{label-standing}{RGB}{148,103,189}
\definecolor{label-sitting} {RGB}{140, 86, 75}
\definecolor{label-lying}   {RGB}{127,127,127}
\definecolor{label-falling} {RGB}{188,189, 34}
\definecolor{label-transit} {RGB}{ 23,190,207}

\IEEEoverridecommandlockouts

\title{\LARGE \bf Multitask Reinforcement Learning for Quadcopter Attitude Stabilization and Tracking using Graph Policy}
\author{Yu Tang Liu$^{1,2,3}$, Afonso Vale$^{3}$, Aamir Ahmad$^{2,1}$, Rodrigo Ventura$^{3}$, Meysam Basiri$^{3}$.	
\thanks{$^1$Max Planck Institute for Intelligent Systems, 72076 T{\"u}bingen, Germany. $^2$University of Stuttgart, 70569 Stuttgart, Germany. $^3$Instituto Superior Técnico, Av. Rovisco Pais 1, 1049-001 Lisboa, Portugal. \texttt{yutang.liu@tuebingen.mpg.de, aamir.ahmad@ifr.uni-stuttgart.de, (afonsocorreiavale, rodrigo.ventura, meysam.basiri)@tecnico.ulisboa.pt}
This work was supported by Aero.Next project (PRR - C645727867- 00000066) and LARSyS FCT funding (DOI: 10.54499/LA/P/0083/2020, 10.54499/UIDP/50009/2020 and 10.54499/UIDB/50009/2020. 
}
}

\makeatletter
\let\NAT@parse\undefined
\makeatother

\usepackage[backend=biber,style=numeric,sorting=none]{biblatex}
\usepackage[english]{babel}
\usepackage{hyperref}

\begin{document}

	\maketitle 

	\makeatletter

\makeatother
\begin{abstract}
Quadcopter attitude control involves two tasks: smooth attitude tracking and aggressive stabilization from arbitrary states. Although both can be formulated as tracking problems, their distinct state spaces and control strategies complicate a unified reward function. We propose a multitask deep reinforcement learning framework that leverages parallel simulation with IsaacGym and a Graph Convolutional Network (GCN) policy to address both tasks effectively. Our multitask Soft Actor-Critic (SAC) approach achieves faster, more reliable learning and higher sample efficiency than single-task methods. We validate its real-world applicability by deploying the learned policy—a compact two-layer network with 24 neurons per layer—on a Pixhawk flight controller, achieving 400 Hz control without extra computational resources. We provide our code at \url{https://github.com/robot-perception-group/GraphMTSAC\_UAV/}.
\end{abstract}



\setlength\belowcaptionskip{-6pt} 

\section{Introduction}
\label{sec:Introduction}
Deep reinforcement learning (DRL) has demonstrated notable success in robotics, including autonomous driving \cite{kiran2021deep} and drone racing \cite{kaufmann2023champion}. However, DRL often struggles by low sample efficiency, unstable training, and limited adaptability—requiring costly retraining or fine-tuning as task or environment conditions change \cite{zhao2020sim, liu2024task}. In contrast, quadcopter attitude control involves two distinct tasks—smooth attitude tracking and aggressive stabilization from arbitrary states—that require different control strategies. This divergence complicates the design of a reward function for single-task RL methods; for example, smooth tracking benefits from penalizing non-smooth or aggressive actions whereas aggressive stabilization does not.

To address these challenges, we propose a multitask DRL framework that relies on parallel simulation. Compare to single task learning, multitask approach offers three key advantages: (1) improved exploration, (2) progressive learning, and (3) enhanced multitask capability. First, by exposing the agent to a variety of tasks, the framework encourages exploration across different regions of the state space, thereby increasing the likelihood of discovering high-reward areas or novel solutions \cite{riedmiller2018learning}. Second, many sparse-reward tasks are easier to solve when simpler skills have already been mastered; for example, learning to run is more feasible once walking has been developed \cite{li2024reinforcement}. Finally, a multitask model retains multiple skills simultaneously, which provides superior adaptability when task requirements change. In contrast, a single-task agent may lose proficiency in previously learned behaviors, necessitating further fine-tuning \cite{liu2024task}.

\begin{figure*}[t]
	\begin{subfigure}[b]{.18\linewidth}
		\includegraphics[width=\linewidth]{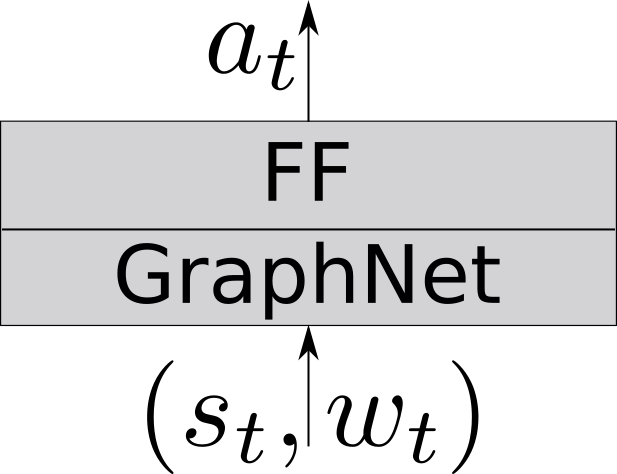}
		\caption{Graph Network \red{(Ours)}}
		\label{fig:architectgraph}
	\end{subfigure}
	\hfill
	\begin{subfigure}[b]{.18\linewidth}
		\includegraphics[width=\linewidth]{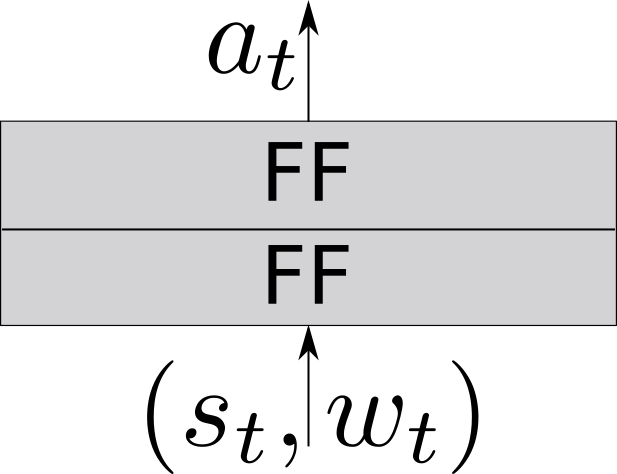}
		\caption{Feed-Forward Network}
		\label{fig:architectplain}
	\end{subfigure}
	\hfill
	\begin{subfigure}[b]{.25\linewidth}
		\includegraphics[width=\linewidth]{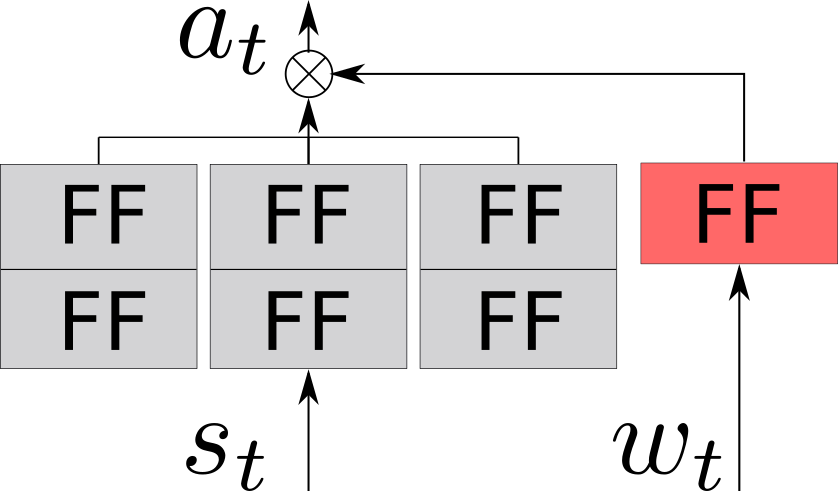}
		\caption{Parallel Network}
		\label{fig:architectparellel}
	\end{subfigure}
	\hfill
	\begin{subfigure}[b]{.27\linewidth}
		\includegraphics[width=\linewidth]{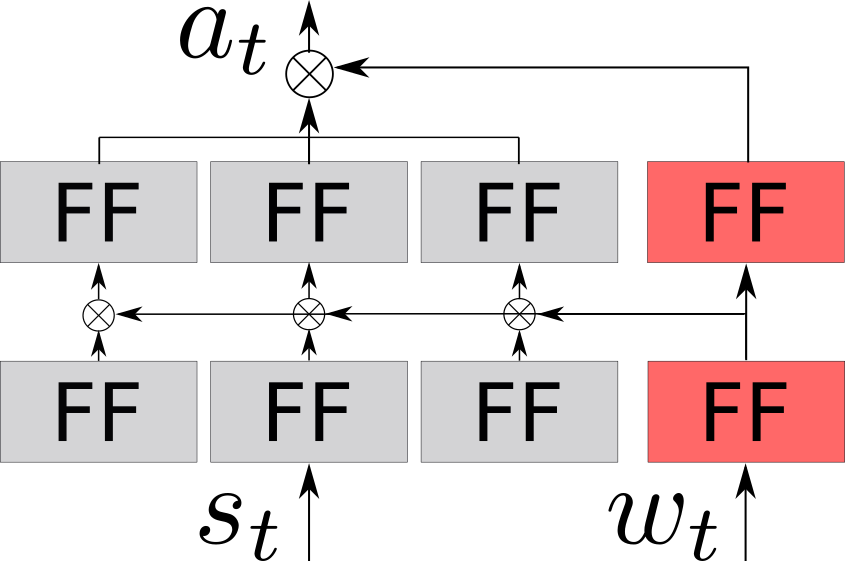}
		\caption{Compositional Network}
		\label{fig:architectcomposition}
	\end{subfigure}
	\caption{Existing Multitask Policy Architectures. The notation $s_t,w_t,a_t$ stands for state, task, and action at time step $t$ respectively. The $FF$ stand for standard feed-forward network with activation function. The red modules are known as context encoder. There exist variations for the context encoder and designing one is in general non-trivial. For simplicity, we consider only the encoder with linear layers as our baselines.}
	\label{fig:policyArchitectures}
\end{figure*}

Despite these advantages, multitask reinforcement learning faces several obstacles, notably in designing an architecture that can effectively retain and manage multiple skills. Ideally, the model should maintain distinct functional sub-modules that can be selectively activated based on the task context \cite{sodhani2021multi, yang2020multi, NEURIPS2022_86b8ad66, iyer2022avoiding} (Fig.~\ref{fig:policyArchitectures}(c,d)). In practice, training such networks is often unstable and yields lower success rates compared to single-task agents. Even slight changes in task requirements may lead to significant rearrangements of these sub-modules, resulting in entirely different behaviors. This issue is further exacerbated by conflicting gradients \cite{yu2020gradient}, where noisy or conflicting gradients reduce learning efficiency. Various approaches have been proposed to mitigate these issues, including architectures that better encode task context or sub-module representation \cite{sodhani2021multi} and introducing sparsity so that only a subset of the network is activated during a given task \cite{iyer2022avoiding}, thereby reducing gradient interference.

To further address these challenges, we introduce a Graph Convolutional Network (GCN) policy \cite{zhang2019graph} that explicitly models the relationships among states, tasks, and actions by carefully tuning the initial adjacency matrix. This explicit modeling encodes prior insights into inter-node relationships—such as decoupling the pitch angle from roll control by setting the corresponding edge weight to zero, or reducing the influence of yaw control on roll and pitch—thereby mitigating gradient conflicts. By blocking gradient flow on irrelevant edges and allowing the adjacency matrix to be tuned during training, our framework enables the agent to discover new interdependencies and achieve faster, more robust learning.

We validate our framework on low-level quadcopter attitude control tasks—specifically, attitude tracking and stabilization—which require distinct control strategies (smooth adjustments versus aggressive maneuvers). Our experiments, conducted using IsaacGym \cite{makoviychuk2021isaac} with domain randomization, demonstrate that our GCN policy trained with multitask Soft Actor-Critic (SAC) outperforms conventional methods. In real-world tests on an F450 quadcopter, our compact GCN model (comprising two layers with 24 neurons each, 697 parameters in total) runs at $\sim400$~Hz entirely on the Pixhawk-6X flight controller without external computational resources, successfully handling both tracking and stabilization under challenging conditions.

Our contributions are summarized as follows:
\begin{itemize}
	\item We propose a novel GCN-based architecture for multitask DRL that leverages domain knowledge to enhance exploration and sample efficiency.
	\item We develop a realistic UAV simulation environment using IsaacGym \cite{liang2018gpu} with tensorized dynamics and a RotorS-based \cite{Furrer2016} rotor model.
	\item We design a compact model optimized for onboard deployment, achieving high-frequency control without requiring GPUs.
	\item We validate our approach in real-world experiments on a Pixhawk-6X F450 quadcopter, demonstrating robust attitude tracking, stabilization, and resilience in extreme drop tests.
\end{itemize} 

\section{Related Work}
\label{sec:2_related}

\textbf{UAV Controller Design.} Traditional UAV controllers like PID \cite{wang2016dynamics} are simple but require extensive tuning and struggle beyond linear regimes, while model-based methods offer better performance at the cost of complex calibration under varying aerodynamic conditions. Recently, DRL has emerged as a flexible alternative for handling high-dimensional inputs and non-convex reward functions, enabling tasks such as drone racing \cite{kaufmann2023champion} and adapting to different quadcopters \cite{zhang2024learning}. Advances in multitask RL—ranging from transfer learning for blimps \cite{liu2024task} to multi-critic architectures for quadcopters \cite{xing2024multi}—further motivate our work, which builds on these developments by introducing a graph-based architecture that supports multitask learning while reducing network size.

\textbf{Multitask RL.} Multitask reinforcement learning seeks to solve multiple tasks simultaneously via shared representations \cite{d2024sharing}. Approaches include meta-RL \cite{wang2016learning}, hierarchical methods \cite{pateria2021hierarchical}, and distillation techniques \cite{ghosh2017divide}, as well as modular architectures using mixture-of-experts or compositional networks \cite{sodhani2021multi, hendawy2023multi, lan2024contrastive, yang2020multi, NEURIPS2022_86b8ad66, iyer2022avoiding} (see Fig.~\ref{fig:policyArchitectures}). However, these methods can be sensitive to hyper-parameters and thus weakening training stability. In contrast, our GCN-based architecture embeds domain knowledge directly into a compact network, thereby boosting learning speed and training success.

\textbf{UAV Simulation Platforms.} Various UAV simulation platforms exist: Flightmare \cite{song2021flightmare} supports up to 300 parallel quadcopter simulations; Gazebo \cite{Furrer2016} offers realistic motor dynamics with ROS integration; and commercial tools like X-Plane \cite{garcia2010multi}, RealFlight, and AirSim \cite{shah2018airsim} provide detailed dynamics and photorealistic rendering. Our framework, built on IsaacGym \cite{makoviychuk2021isaac}, enables over 1,000 parallel simulations on a single GPU, significantly enhancing training speed. We further augment the simulation by integrating a RotorS-based \cite{Furrer2016} motor dynamics model, including air drag, rolling moment, etc., in PyTorch, ensuring seamless incorporation into our RL training loop.

%

\setlength{\belowdisplayskip}{0pt} \setlength{\belowdisplayshortskip}{0pt}
\setlength{\abovedisplayskip}{0pt} \setlength{\abovedisplayshortskip}{0pt}

\setlength\belowcaptionskip{0pt} 

\section{Preliminary}
\label{sec:Preliminary}

\begin{figure*}
	\centering
	\includegraphics[width=0.95\linewidth]{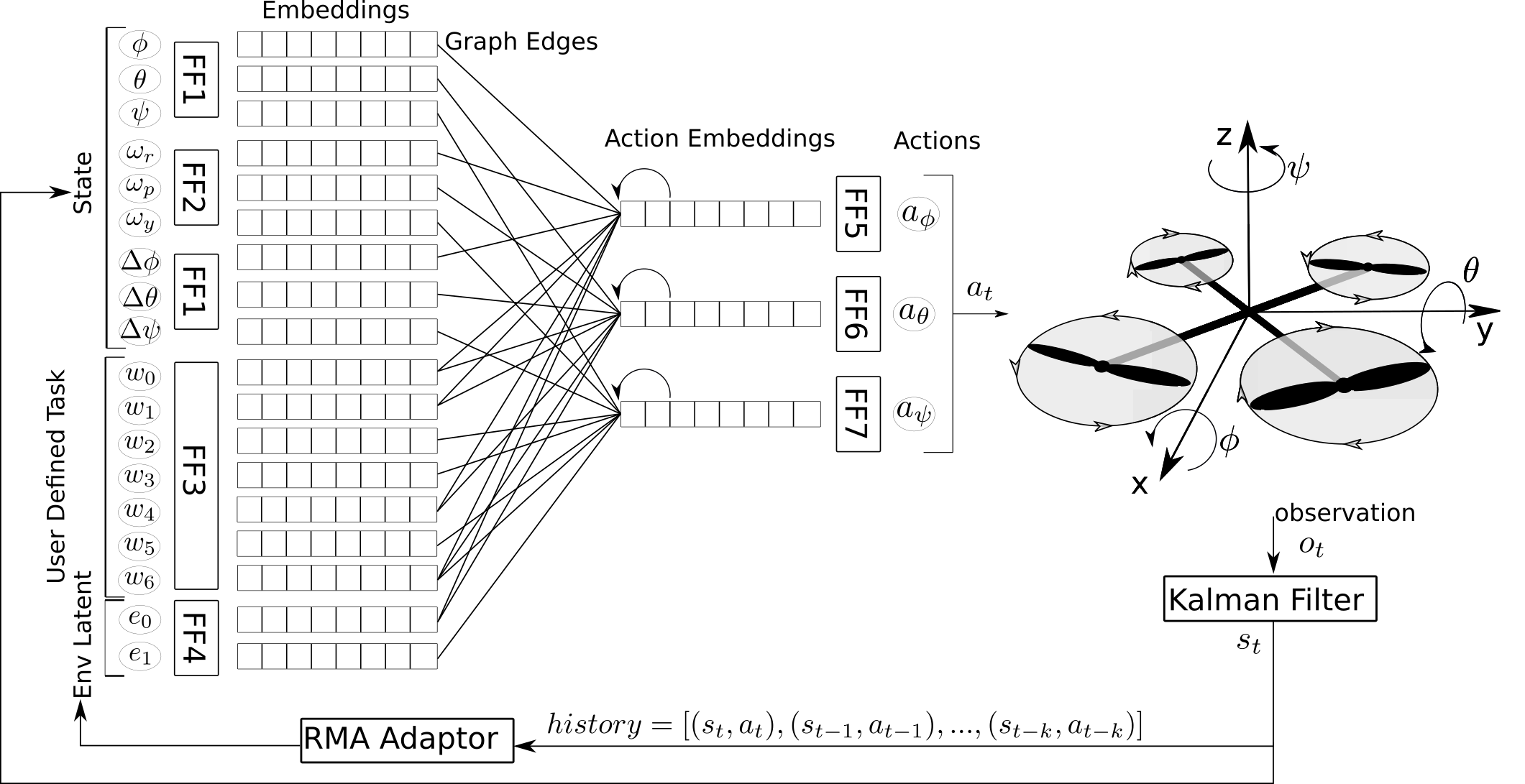}
	\caption{Illustration of our GCN-based policy.The policy observes the robot state $s_t$, the task weight $w_t$, and the environment latent variables $e_t$. These inputs are first passed through feed-forward layers $\text{FFX}$ to produce node embeddings. The GCN then updates embeddings by multiplying with a linear layer $W^{(l)}$ (not shown) and an adjacency matrix $A$ constructed by stacking the graph edges, resulting in action-node embeddings. Each action node has a recurrent edge to itself, incorporating the previous action. Finally, we project action embeddings to scalar values for the control commands. A Kalman filter provides state estimates, and a history buffer feeds the RMA adapter.}
	\label{fig:graphnet}
\end{figure*}

\subsection{Markov Decision Process and Multitask Reinforcement Learning} 
\textit{RL} improves a policy through interaction with the environment, formulated as a \textit{Markov Decision Process} (MDP), $\mathcal{M}\equiv\langle \mathcal{S}, \mathcal{A}, p, r, \gamma \rangle$. Here, $\mathcal{S}$ is the state space, and $\mathcal{A}$ is the action space. The transition function $p(s_{t+1} \mid s_t, a_t)$ describes the probability of moving from one state to another under an action, while the reward function $r(s_t, a_t, s_{t+1})$ guides the agent’s learning signal. The discount factor $\gamma \in [0,1)$ sets the time horizon. The objective is to find an optimal control policy $\pi(a \mid s): \mathcal{S}\rightarrow \mathcal{A}$ that maximizes the expected discounted return,
\begin{align} \label{eqn:mpd_return}
	J(\pi)=\mathbb{E}_{\pi, \mathcal{M}}[\sum_{\tau=t}^{\infty} \gamma^{\tau-t}r_\tau].
\end{align}

\textit{Multitask RL} extends this setup to a family of MDPs $\mathcal{M}_w$, each characterized by a different reward function: $\mathcal{M}_w \equiv<\mathcal{S},\mathcal{A},p,r_w,\gamma>$, where $r_w(s_t, a_t, s_{t+1})=\phi(s_t, a_t, s_{t+1})^\top \cdot w$. Here, $\phi(s_t, a_t,s_{t+1})\in \mathbb{R}^d$ is a feature vector shared among tasks, and $w \in \mathbb{R}^d$ encodes the task-specific weighting of these features. The goal is to find a single policy that performs well over the task space $\mathcal{W}$ or a desired task $w_{desired} \in \mathcal{W}$ with $w_{desired}$ being an unknown vector.

\subsection{Quadcopter State and Action Space}
Our framework focuses on attitude control, so the state vector is defined as
\begin{equation}
	s = \begin{bmatrix}
		\phi & \theta & \psi & \omega_r & \omega_p & \omega_y & \Delta \phi & \Delta \theta & \Delta \psi
	\end{bmatrix}^\top,
\end{equation}
where \(\phi\), \(\theta\), and \(\psi\) are the roll, pitch, and yaw angles, \(\omega_r\), \(\omega_p\), and \(\omega_y\) denote their corresponding angular velocities, and \(\Delta \text{Angle} \) denote the angular error between angle and the target angle. The action vector comprises low-level attitude control commands:
\begin{equation}
	a = \begin{bmatrix}
		a_\phi & a_\theta & a_\psi
	\end{bmatrix}^\top,
\end{equation}
which are mapped to rotor thrust adjustments to control the UAV's attitude. The altitude control and target attitude are both provided by ArduPilot’s built-in Cascade PID controllers.

\begin{figure}[H]
	\centering
	\includegraphics[width=0.95\linewidth]{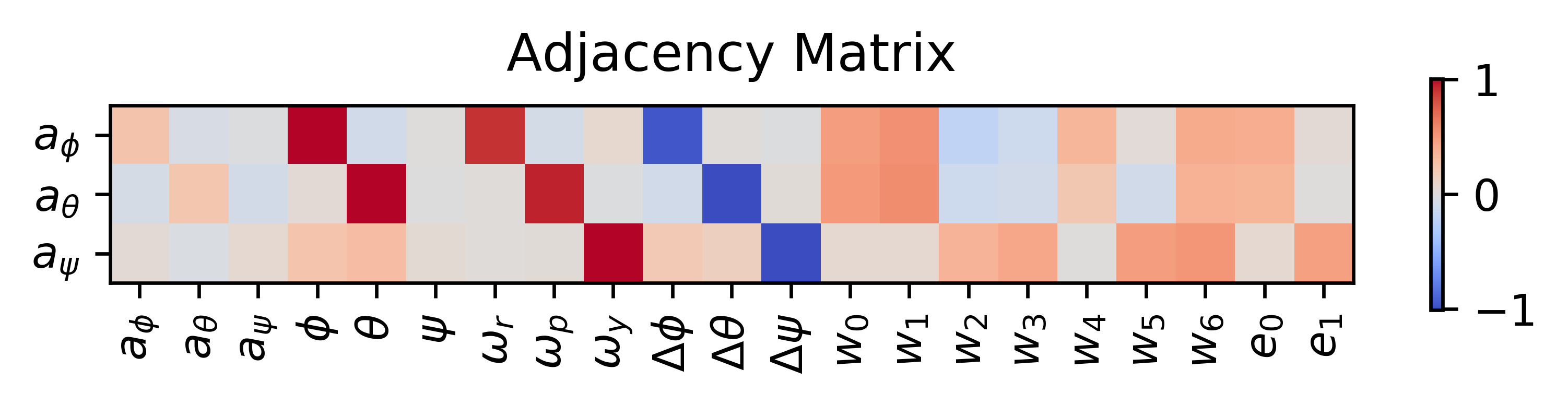}
	\caption{GCN weight visualization. Each row represents a different action dimension (\emph{e.g.,} roll, pitch, yaw), and each column corresponds to an input dimension from the state or task embeddings. Darker cells indicate stronger learned correlations.}
	\label{fig:exp3graph}
\end{figure}

\section{Methodology}
\label{sec:Methodology}
Below we will introduce our Quadcopter multitask simulation based on IsaacGym \cite{makoviychuk2021isaac} as well as the graph policy trained using our multitask SAC.

\subsection{Feature and Task Formulation}

In our framework, the UAV reward function is defined as a linear combination of these features:
\begin{equation}\label{eqn:rewardfeaturetask}
	r_w(s_t, a_t, s_{t+1}) = \phi(s_t, a_t, s_{t+1})^\top w,
\end{equation}

where each task is represented by a weight vector \(w \in \mathbb{R}^d\) and \(\phi(s_t, a_t, s_{t+1}) \in \mathbb{R}^d\) encapsulates key aspects of UAV performance (e.g., tracking error, angular velocity error, and action smoothness).

\subsubsection{Feature Function Formulation}
We define three auxiliary functions:
\begin{align}
	\mathrm{quad}(d,c) &= \frac{1}{1+c\,d^2},\\[1mm]
	\mathrm{binary}(d,c) &= 
	\begin{cases}
		1, & d < c,\\[1mm]
		0, & \text{otherwise},
	\end{cases}\\[1mm]
	\mathrm{linear}(d,c) &= -c\cdot d.
\end{align}
Next, we define the following distance metrics: 
\begin{itemize}
	\item \(d_1 = \sqrt{(\Delta\phi)^2+(\Delta\theta)^2}\), where \(\Delta\phi = \phi - \phi_{\text{goal}}\) and \(\Delta\theta = \theta - \theta_{\text{goal}}\) (roll and pitch error),
	\item \(d_2 = |\Delta\psi|\) with \(\Delta\psi = \psi - \psi_{\text{goal}}\) (yaw error),
	\item \(d_3 = \sqrt{\omega_r^2+\omega_p^2}\) (roll and pitch rate norm),
	\item \(d_4 = |\omega_y|\) (yaw rate norm),
	\item \(d_5 = \|a^{\text{prev}}_{\phi,\theta,\psi} - a_{\phi,\theta,\psi}\|\) (action smoothness).
\end{itemize}
With these definitions, the feature extraction function \(\phi: \mathcal{S}\times\mathcal{A}\times\mathcal{S}\to\mathbb{R}^7\) is given by
\begin{equation}\label{eqn:phimatrix}
	\phi(s,a,s') = \begin{bmatrix}
		0.5\,\mathrm{quad}(d_1,10) + 0.5\,\mathrm{binary}(d_1,0.1) \\[1mm]
		0.5\,\mathrm{quad}(d_1,100) + 0.5\,\mathrm{binary}(d_1,0.01) \\[1mm]
		0.5\,\mathrm{quad}(d_2,10) + 0.5\,\mathrm{binary}(d_2,0.1) \\[1mm]
		0.5\,\mathrm{quad}(d_2,100) + 0.5\,\mathrm{binary}(d_2,0.01) \\[1mm]
		0.5\,\mathrm{quad}(d_3,10) + 0.5\,\mathrm{binary}(d_3,0.1) \\[1mm]
		0.5\,\mathrm{quad}(d_4,10) + 0.5\,\mathrm{binary}(d_4,0.1) \\[1mm]
		\mathrm{linear}(d_5,1)
	\end{bmatrix}.
\end{equation}
Each feature and its value $c\in\mathbb{R}$ were determined empirically. The first two features promote roll and pitch tracking by encouraging errors below 0.1 and 0.01 rad, respectively; the third and fourth similarly reduce yaw error. The fifth and sixth features penalize high angular velocities, while the final feature promotes action smoothness. In each case, the discrete \(\mathrm{binary}\) term provides a bonus when the error is below \(c\), and the continuous \(\mathrm{quad}\) term reduces reward sparsity to aid convergence.

\subsubsection{Task and Task Set}
A \emph{task} is defined by its weight vector \(w \in \mathbb{R}^d\), and a \emph{task set} is a collection of such vectors:
\begin{equation} 
	\mathcal{W}^i = \{w^i_j \mid j \in \mathcal{N}\}.
\end{equation}
Each task set targets a specific control objective. For example:
\begin{itemize}
	\item Tracking Task Set: These tasks aim to maintain the vehicle near a target (e.g., zero) attitude. An example weight vector is 
	\begin{equation} 
		w_{\text{track}} = [1,\,1,\,1,\,1,\,0,\,0,\,0]^\top,
	\end{equation}
	where the first four components emphasize angle tracking while the remaining entries downweight other criteria.
	\item Smooth Tracking Task Set: These tasks introduce angular velocity penalty and promote control smoothness in addition to accurate attitude tracking. For instance, one weight vector is 
	\begin{equation} 
		w_{\text{smooth}} = [1,\,1,\,1,\,1,\,0.3,\,0.3,\,0.3]^\top.
	\end{equation}
	\item Stabilization Task Set: These tasks allow for arbitrary initial attitudes and focus on restoring a stable configuration. An example is 
	\begin{equation} 
		w_{\text{stab}} = [1,\,0,\,0,\,0,\,0,\,0,\,0]^\top.
	\end{equation}
\end{itemize}
We specified a total number of 57 tasks split in these task sets. To ensure consistent reward scales, each task weight vector is normalized and sampled uniformly per environment. 

\subsection{Multitask SAC Training Scheme}
We adopt the SAC algorithm \cite{haarnoja2018soft}, which alternates between soft policy evaluation and soft policy improvement steps to gradually improve the policy performance by tuning the network parameters. Our multitask variant pools experience from parallel simulations, each initialized with a task vector $w$ sampled from a predefined task set, i.e. $w \sim W^i$. Instead of storing raw rewards in the replay buffer, we store only tuple $(s_t,a_t,s_{t+1})$ . For each mini-batch, we then apply dot product between all task weights and a feature function $\phi (s_t,a_t,s_{t+1} )$ to the same transitions, resulting in several reward signals w.r.t different tasks $ r_w = w \cdot \phi ( s_t,a_t,s_{t+1} )$, enabling efficient sharing of experiences across tasks.

This scheme is particularly beneficial when tasks have overlapping objectives. For example, the tracking set has the initial condition as the goal of the stabilization set. Thus allows agent to explore high reward region for the stabilization tasksets. 

\subsection{Graph Neural Network for Continuous Control}
Figure~\ref{fig:graphnet} provides an overview of our GCN-based policy architecture. The agent receives (i) the current robot state $s_t$, (ii) the task vector $w_t$, (iii) an environment latent variable $e_t$ provided by RMA adaptor \cite{kumar2021rma}. 

Each input is projected into a node embedding, which is then updated via a graph convolution. The action nodes also maintain cyclical connections to incorporate previous control outputs. By explicitly specifying which nodes are connected, the GCN encodes domain-specific priors. As training proceeds, the GCN’s learned adjacency matrices can reveal how each input node influences action nodes, as visualized in Figure~\ref{fig:exp3graph}.
Below we provide the formulation of the GCN policy:

\subsubsection{Input Embedding}:
The network receives two primary inputs: a state vector and a task vector. These vectors are grouped into several modalities. Each modality is independently projected from scalar values to a higher-dimensional embedding via linear layers, denoted by $\text{FFX}$ in the Fig.~\ref{fig:graphnet}. Formally, if $x \in \mathbb{R}^{n}$ is a vector from one modality and $x_i \in \mathbb{R}$ is one of its element, then the embedding of the $k$-th modality is given by:
$h_i=\text{FFk}(x_i) \in \mathbb{R}^{d_{embed}}$. The outputs from all modalities are then concatenated, forming a matrix $H \in \mathbb{R}^{B\times N \times d_{embed}}$, where $B$ is the batch size and $N$ is the total number of nodes (including state, task, and action nodes).

\subsubsection{Graph Construction and Convolution}:
A learnable state-action adjacency matrix $A \in \mathbb{R}^{N \times N}$ is introduced as a parameter of the network. This matrix defines the connectivity between different nodes in the graph, effectively encoding prior knowledge (e.g., preventing undesired connections between non-related nodes). The network then applies a series of GCN layers. Each GCN layer updates the node embeddings by aggregating information from connected nodes:
$H^{(l+1)}=\sigma(\tilde{A}H^{(l)}W^{(l)})$,
where $H^{(l)}$ is the embedding at layer $l$, $\tilde{A}$ is the adjacency matrix clipped in range $[-1,1]$, $W^{(l)}$ is the layer’s weight matrix, and $\sigma(\cdot)$ is a non-linear activation function. In our implementation, the number of GCN layers is configurable, but we have only tested with one layer as number of computation is strictly limited in our setup.

\subsubsection{Action Prediction}:
After the graph convolution steps, the node embeddings corresponding to the action nodes are isolated. Each action embedding is then decoded and produce the mean $\mu$ and the standard deviation $\sigma$ of a Gaussian policy for sampling the action, i.e. $a_t \sim \mathcal{N}(\mu_t, \sigma_t)$.

\begin{figure*}
	\centering
	\begin{subfigure}[t]{0.32\textwidth}
		\centering
		\includegraphics[width=0.95\linewidth]{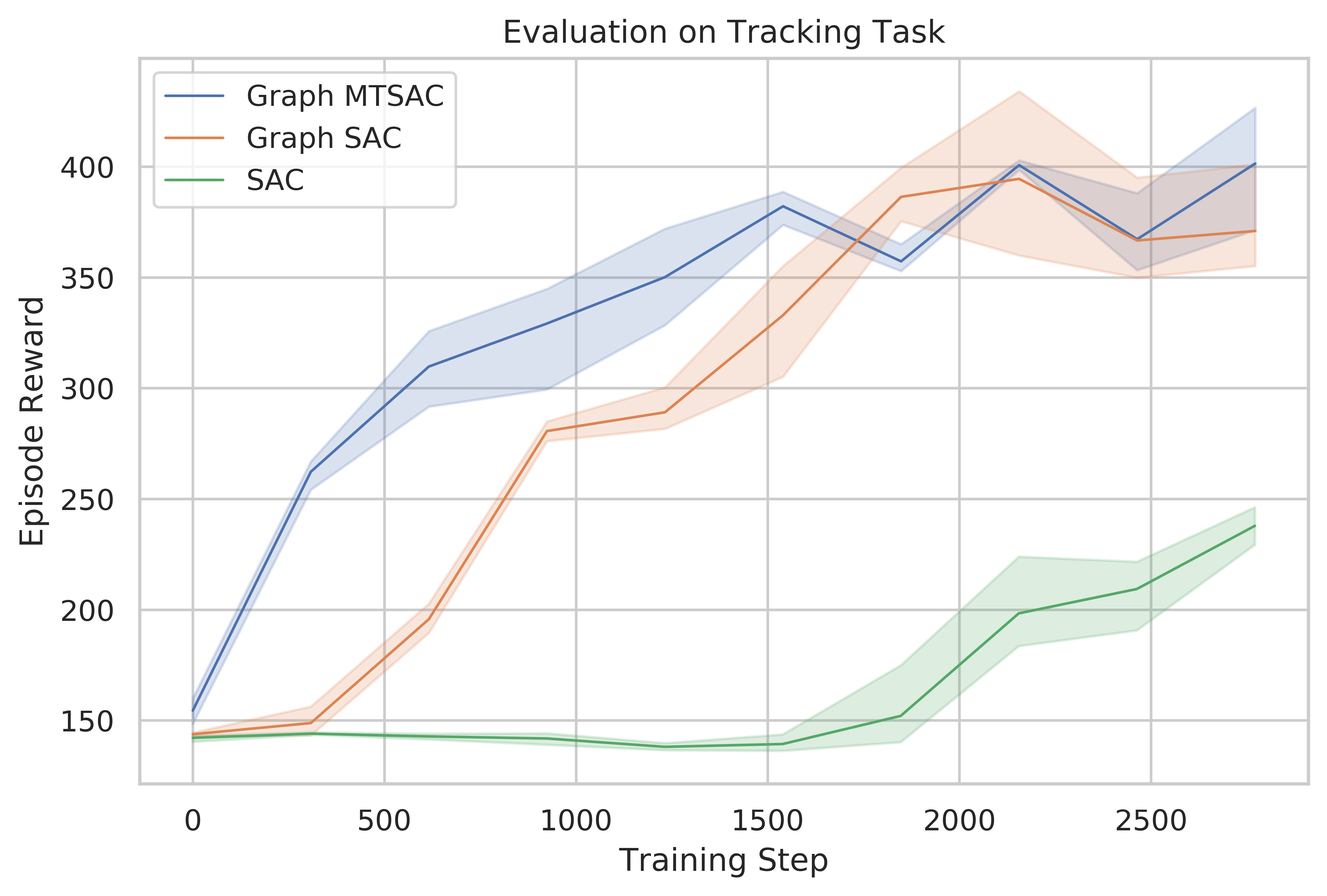}  
		\caption{Multitask vs. Single-Task Training: Graph MTSAC is trained using a diverse set of tasks and evaluated on the angle tracking task, whereas single-task agents are both trained and evaluated solely on angle tracking.}
		\label{fig:exp1singletask}
	\end{subfigure}
	\hfill
	\begin{subfigure}[t]{0.32\textwidth}
		\centering
		\includegraphics[width=0.95\linewidth]{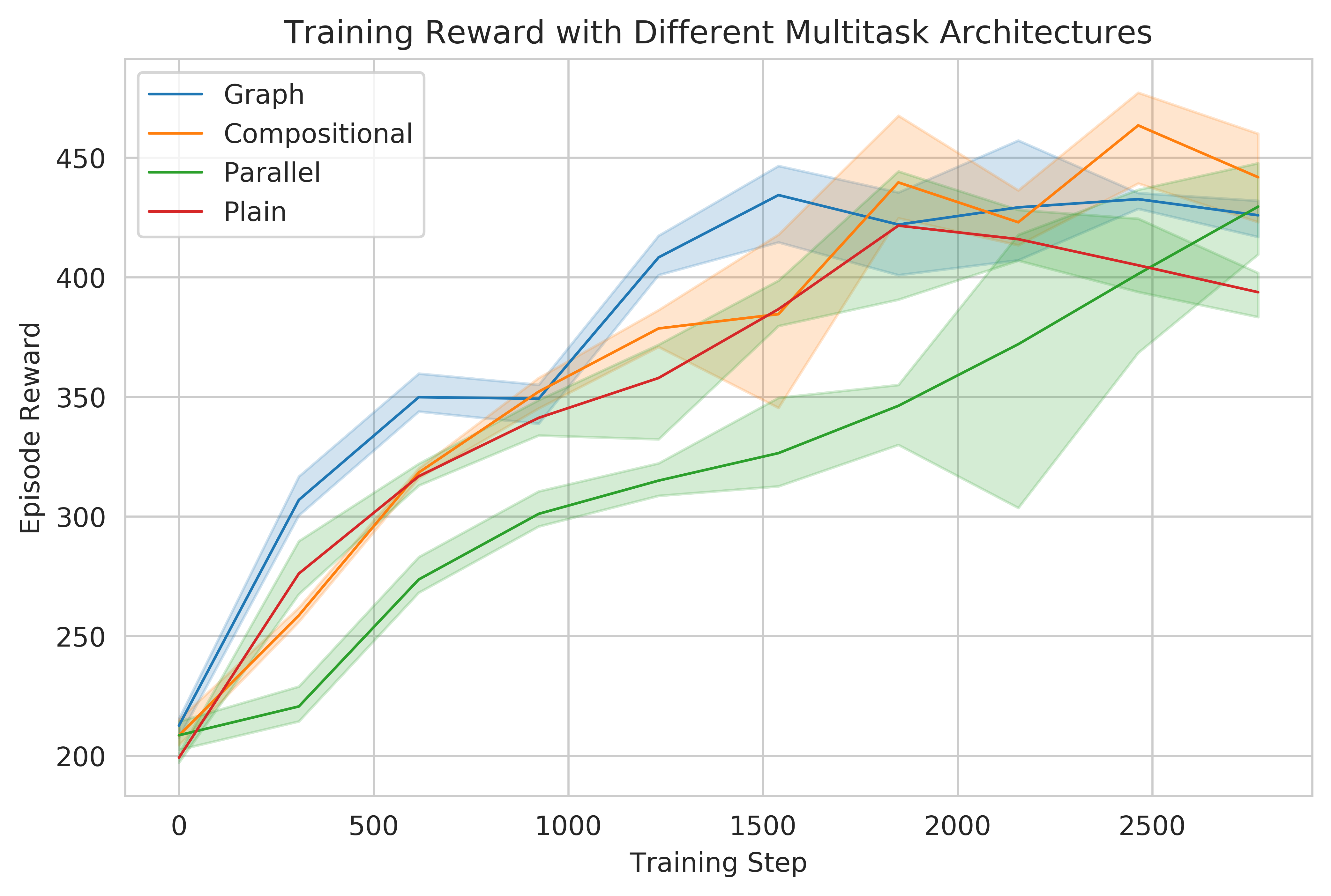}
		\caption{Model Architecture Comparison: Performance of multitask learning across different network architectures. While our graph policy demonstrate higher asymptotic performance, other models remain competitive in multitask scenario.}
		\label{fig:exp3architecture}
	\end{subfigure}
	\hfill
	\begin{subfigure}[t]{0.32\textwidth}
		\centering
		\includegraphics[width=0.95\linewidth]{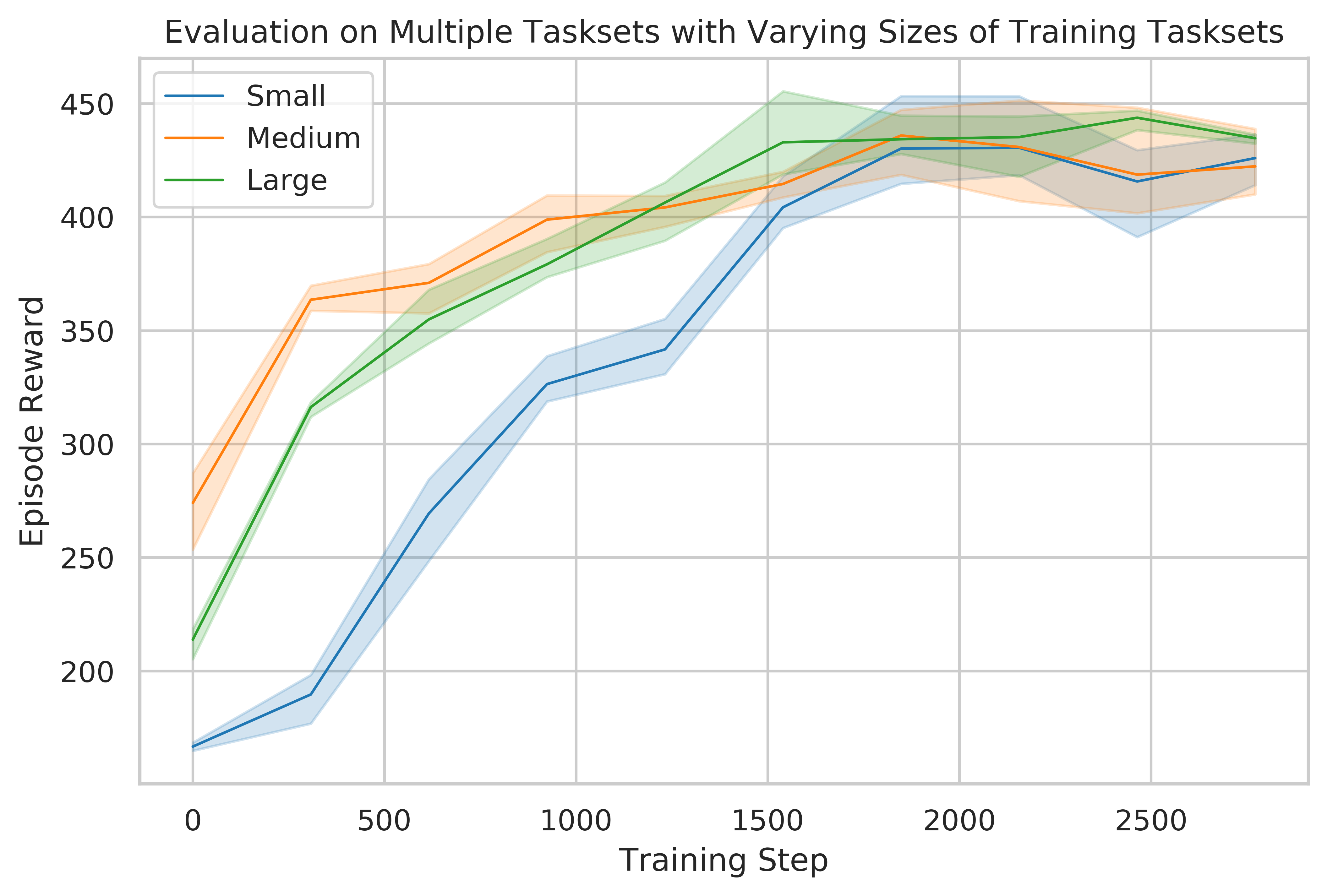}
		\caption{Task Set Diversity: Performance improves with increased task diversity. The small set includes only tracking tasks; the medium set mirrors the evaluation tasks (tracking and stabilization); and the large set adds tasks emphasizing action smoothness.}
		\label{fig:exp4trainsets}
	\end{subfigure}
	\hfill
	\begin{subfigure}[t]{0.32\textwidth}
		\centering
		\includegraphics[width=0.95\linewidth]{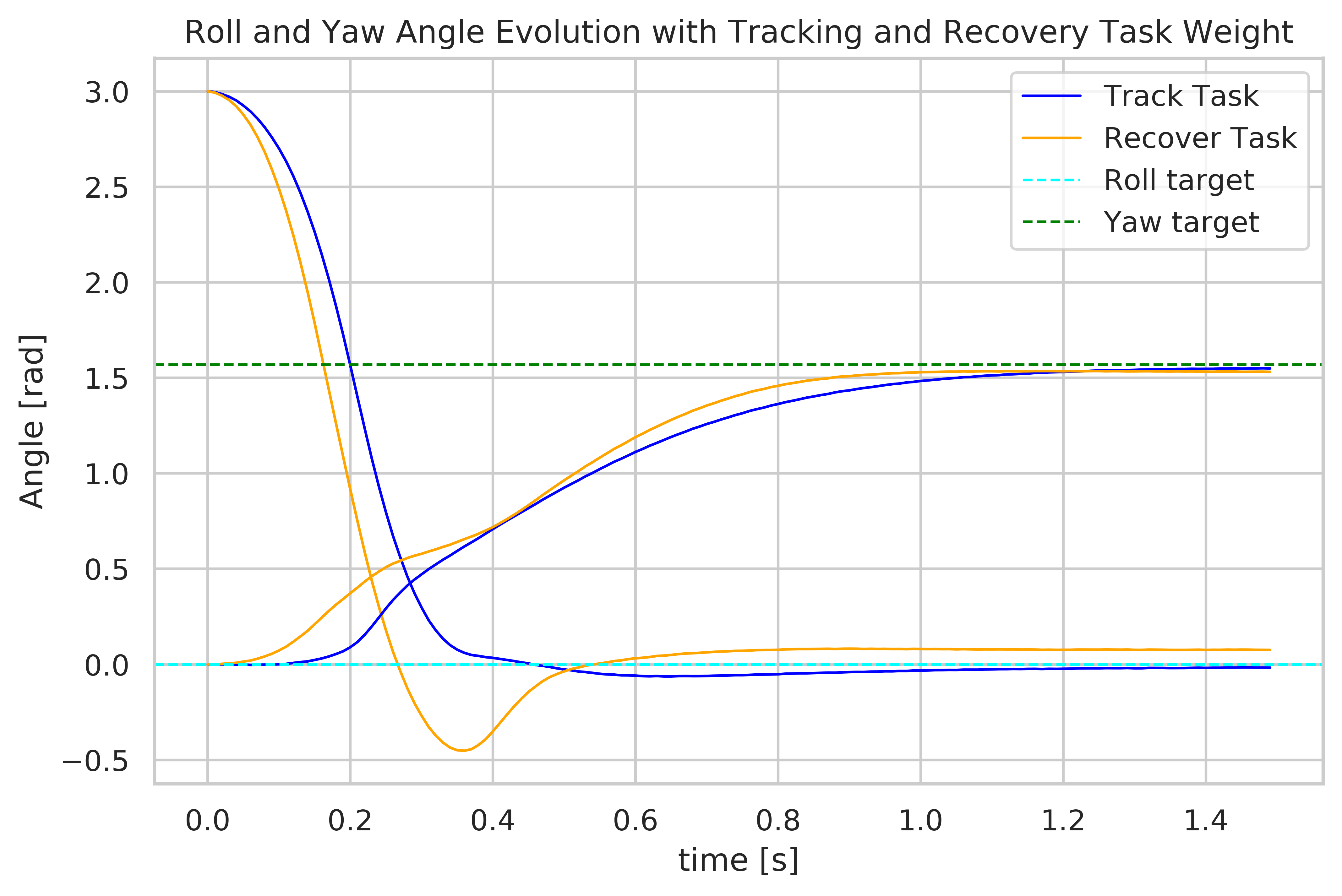}
		\caption{Behavior Modulation: With an initial roll 3~rad and yaw 0~rad, the agent must reach 0 roll and 1.57 yaw. Under a stabilization task weight, the agent exhibits aggressive control; under a tracking task weight, it produces smoother control to avoid overshooting.}
		\label{fig:exp5multitaskproperty}
	\end{subfigure}
	\hfill
	\begin{subfigure}[t]{0.32\textwidth}
		\centering
		\includegraphics[width=0.95\linewidth]{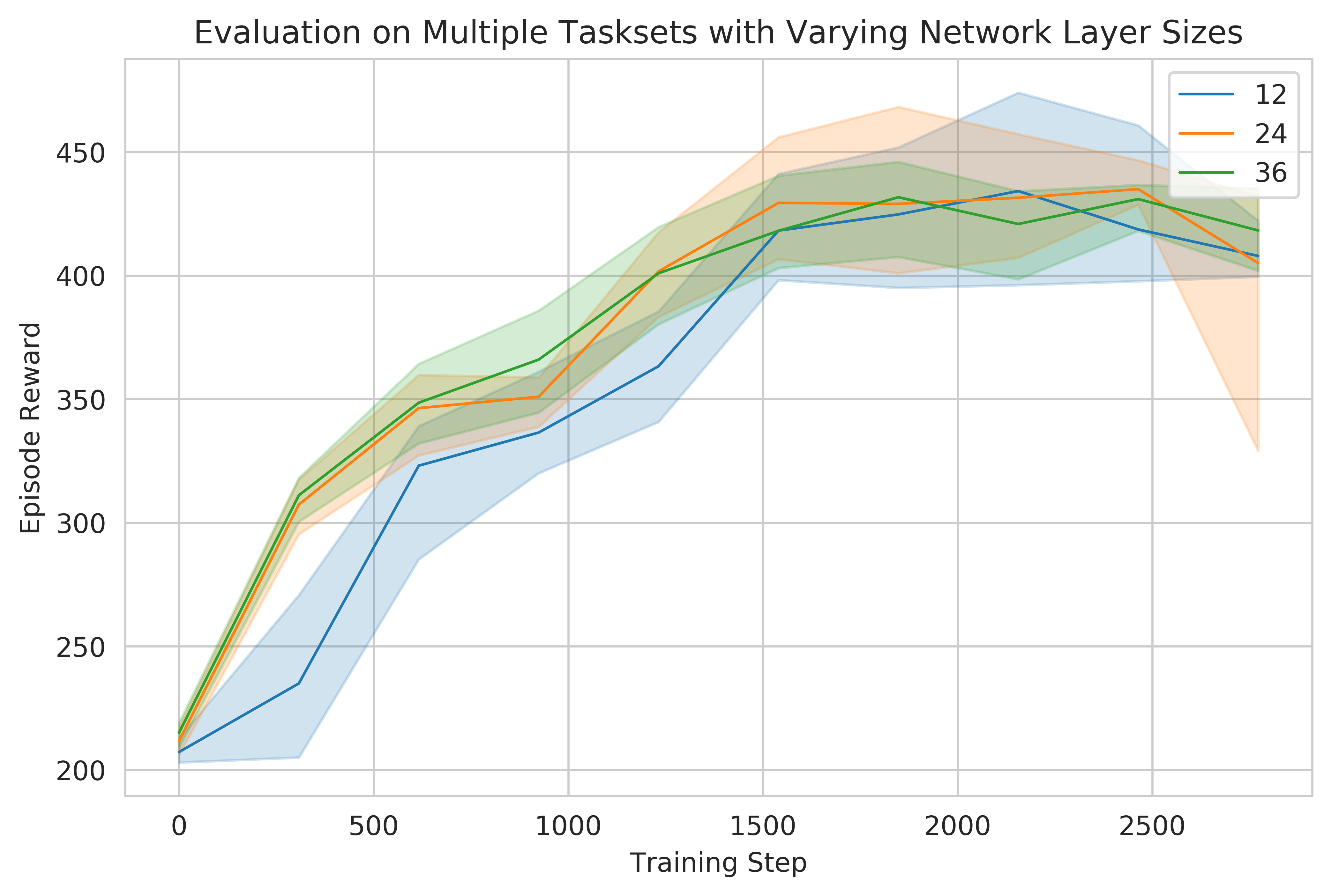}
		\caption{Effect of Network Size: Increasing neurons per layer from 12 to 24 improves sample efficiency and accelerates learning, while further increasing to 36 neurons offers no additional benefit.}
		\label{fig:exp6graphsacnetsize}
	\end{subfigure}
	\hfill
	\begin{subfigure}[t]{0.32\textwidth}
		\centering
		\includegraphics[width=0.95\linewidth]{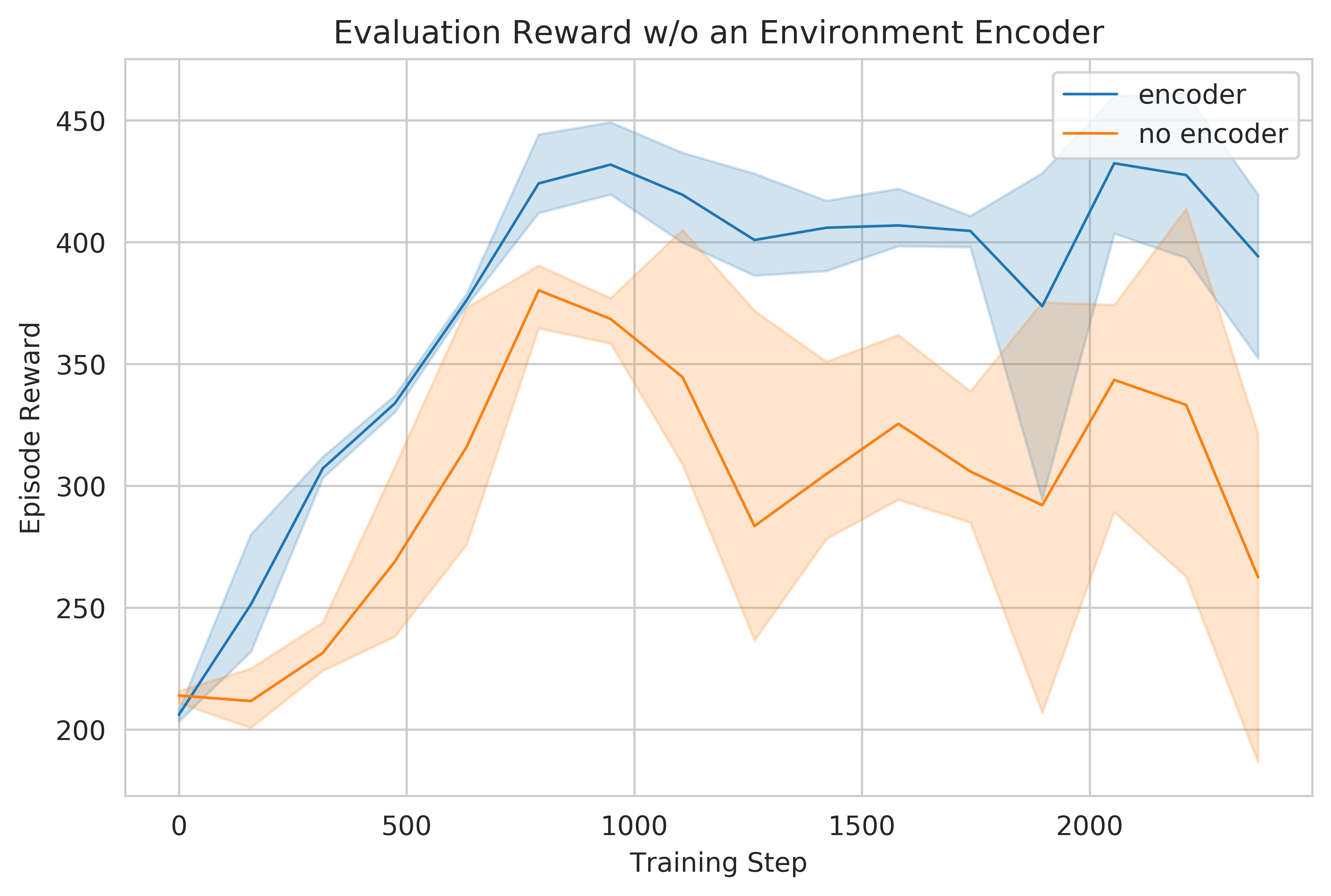}
		\caption{Domain Adaptation: We compare performance with and without an RMA encoder during domain randomized evaluation. Without privileged information about environment factors training becomes unstable.}
		\label{fig:exp7domainadaptation}
	\end{subfigure}
	\caption{Simulation results. Unless otherwise specified, the default network configuration uses 24 neurons per layer. Each experiment was repeated three times with different seeds; the main curve represents the mean performance, and the shaded area indicates the range between the maximum and minimum values.}
	\label{fig:simulationresult}
\end{figure*}

\subsubsection{Discussion of GCN Policy}
The GCN policy offers a structured approach to fusing multi-modal information, which enhances sample efficiency and improves multitask learning by explicitly encoding initial relationships among state, task, and action inputs. This architecture facilitates faster convergence and robust performance in complex control tasks, while remaining compact enough for deployment on resource-constrained devices.

However, the benefits come with trade-offs. The performance is sensitive to the design of the state-action adjacency matrix, requiring careful tuning and domain expertise. Additionally, the added complexity of graph operations may introduce computational overhead and limit flexibility compared to simpler feedforward networks, particularly in scenarios where the optimal relationships among modalities are less structured.

\subsection{Sim-to-Real Transfer via RMA} 
\label{sec:rma} 
We apply RMA \cite{kumar2021rma} and domain randomization \cite{tobin2017domain} (Table.\ref{tab: env latents}) to enhance generalization and facilitate sim-to-real transfer. The method proceeds in three stages:

\begin{enumerate} 
	\item Training Encoder and Policy: Each simulation exposes the environment factor $\alpha \in \mathbb{R}^{\norm{\alpha}}$ such as thrust, torque, or air drag coefficients to the RL agent. An encoder $f^{encoder}(\cdot)$ converts the environment factors $\alpha$ to a latent vector $e$. The policy $\pi$ then takes as input $(s_t, e_t)$ and outputs actions $a_t$. 
	
	\item Training the Adaptor: Once the encoder and policy are trained, we collect rollouts and record the past trajectory $ \{ (s_{t-i}, a_{t-i}) \}_{i=1}^m $.
	We then train an adaptor network $f^{adaptor}$ that predicts the encoder's output $\hat{e}_t=f^{adaptor}(s_{t-1},a_{t-1},s_{t-2},a_{t-2},...)$. Minimizing the supervised loss $\norm{\hat{e}_t-e_t)}$ ensures the adaptor mimics the encoder without needing information of environment factor at test time. 
	
	\item Deployment: At run time, we discard the encoder and use only the trained adaptor. The adaptor predicts the environment latent state from onboard measurements, letting the original policy operate as if it had direct access to privileged parameters. Due to the limited computation, we assume slow environmental changes and restrict the adaptor to run only at 4hz. 
\end{enumerate}
 

\setlength\belowcaptionskip{-6pt}

\section{Experiments and Results}
\label{sec:4_experiment}
In our experiments, we evaluated our GCN-based multitask SAC framework for UAV control using IsaacGym's large-scale simulation to train and validate two key tasks: attitude tracking and stabilization. Our evaluation focuses on three aspects: (1) Sample Efficiency and Stability: Does our GCN-based policy and multitask learning paradigm improve sample efficiency and training stability compared to alternative architectures or single task paradigm? (2) Task Generalization and Robustness: How well does the policy generalize across different task configurations and hyperparameter settings? (3) Sim-to-Real Transfer: Can the simulation-trained policy successfully transfer to real-world deployment on an actual quadcopter?

\begin{table}[h]
	\centering
	\resizebox{0.28\textwidth}{!}{%
		\begin{tabular}{ c||c }
			\hline
			Type A & $C_T$, $C_Q$, \\
			\hline
			Type B & $C_D$, $C_M$, hover state, $\alpha_{up}$, $\alpha_{down}$\\
			\hline
	\end{tabular}}
	\caption{Environment Factors. Type A variables are randomized between 50\% and 200\% of their nominal values, while Type B variables vary from 80\% to 120\%. The coefficients $C_T$, $C_Q$, $C_D$, and $C_M$ correspond to thrust, torque, air drag, and rolling moment, respectively (see RotorS implementation \cite{Furrer2016} for details). The hover state parameter models the current thrust relative to hover thrust, and $\alpha_{up}$ and $\alpha_{down}$ represent the time delays for increasing and decreasing rotor speed.
		\label{tab: env latents} 
	}
\end{table}

\begin{figure*}
	\centering
	\begin{subfigure}[t]{0.5\textwidth}
		\centering
		\includegraphics[width=0.95\linewidth]{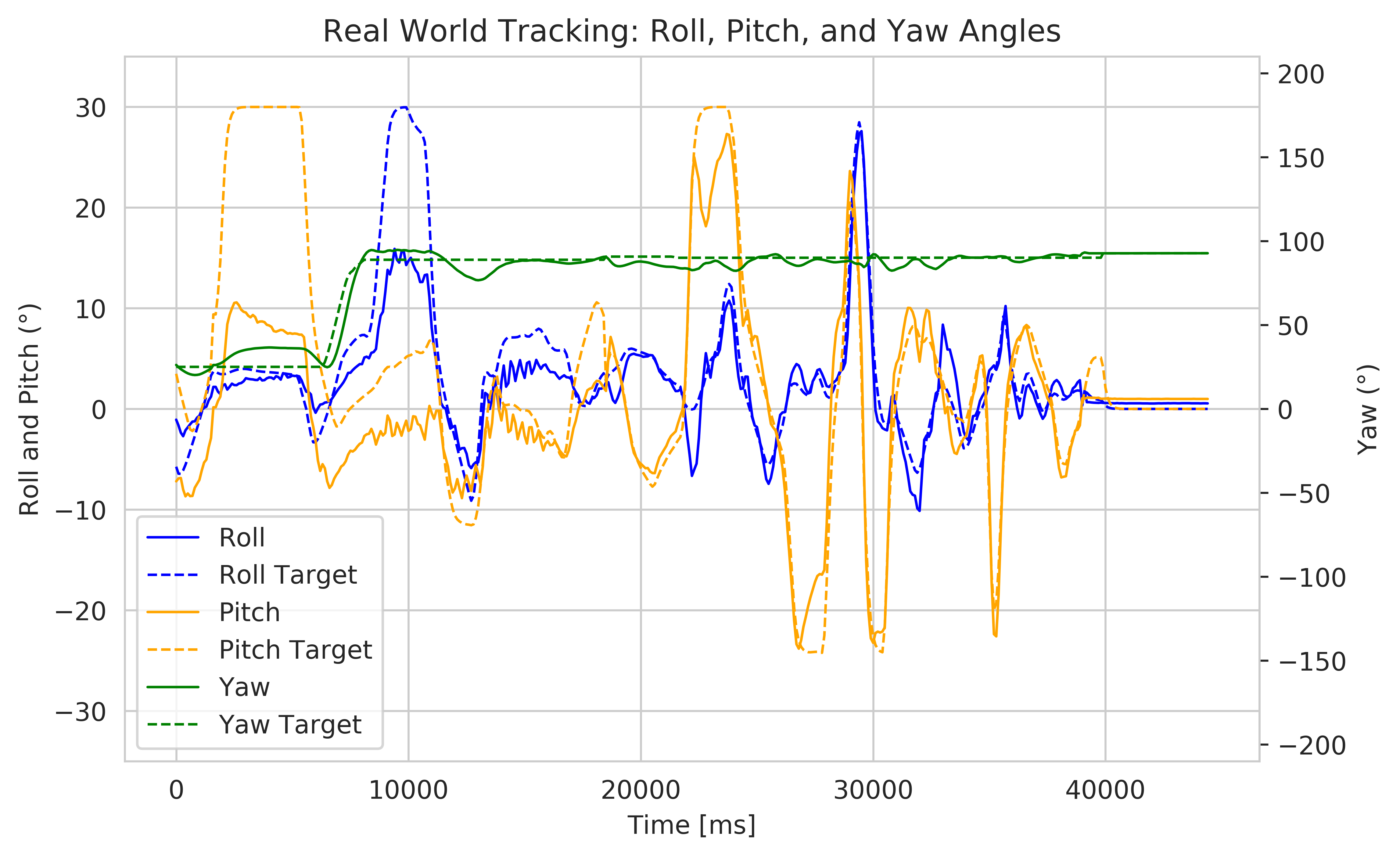}
		\caption{Tracking Task: The task runs in Ardupilot position control mode, where the high-level position PID controller processes pilot's commands and generates low-level attitude commands for the agent. Since the agent was trained with roll and pitch command within a 10-degree range, it accurately follows commands within this range but does not always generalize beyond it.}
		\label{fig:realworldtracking}
	\end{subfigure}
	\hfill
	\begin{subfigure}[t]{0.47\textwidth}
		\centering
		\includegraphics[width=0.95\linewidth]{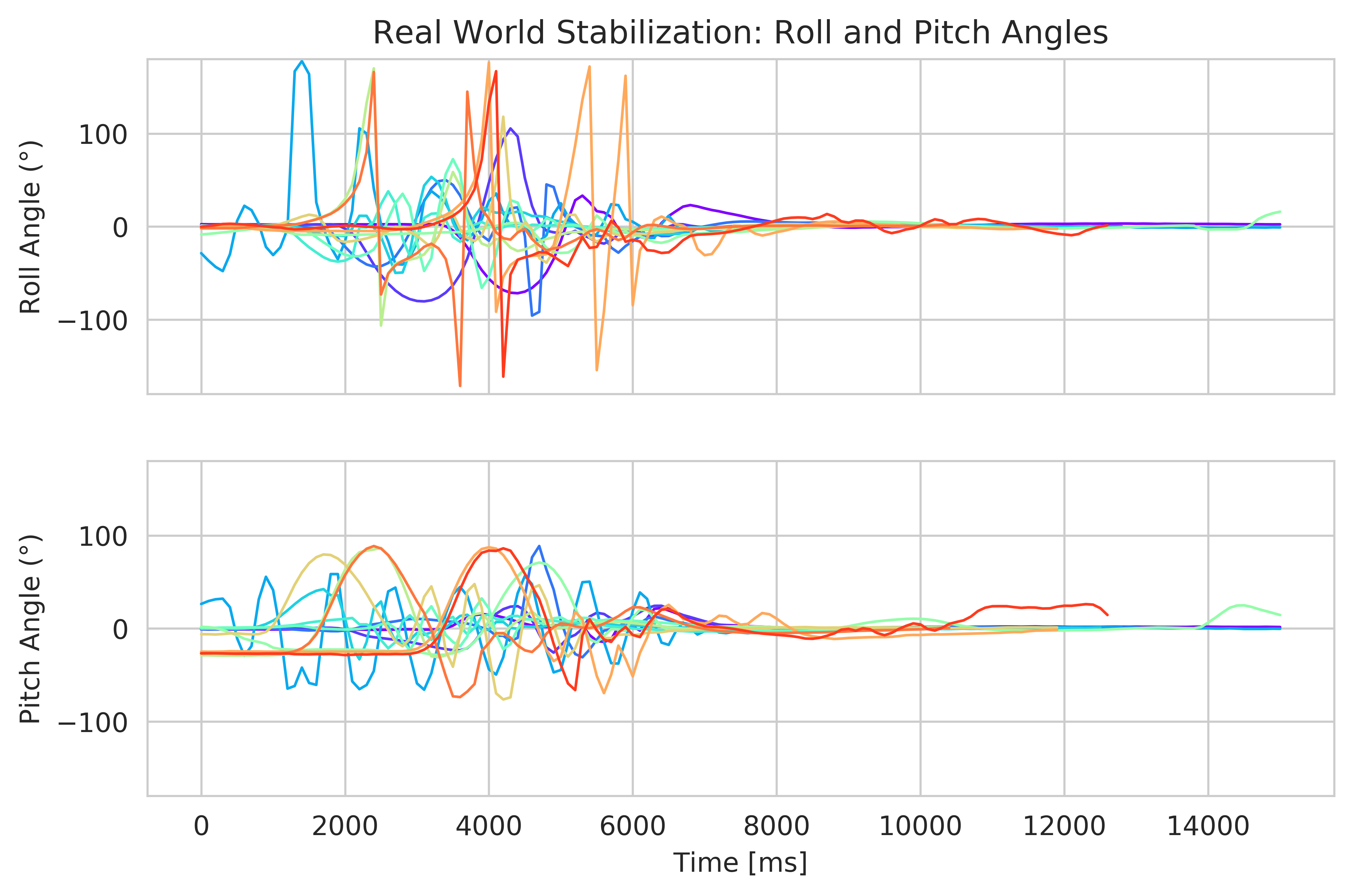}
		\caption{Stabilization Task: After a complete rotor disengagement for 1 to 3 seconds, forcing the quadcopter to enter an unstable free fall, the agent successfully re-engaged the rotors and stabilized the UAV in all 13 trials. This demonstrates the model’s ability to recover from extreme orientations and high angular velocities.  Curves in same color indicate roll and pitch angle of the same experiment.}
		\label{fig:realworldfreefallrollpitch}
	\end{subfigure}
	
	\caption{Real World Result. The tracking task applies the task weight $[1,1,1,1,0,0,0]$ whereas the stabilization task has the task weight $[1,0,0,0,0,0,0]$.} 
	\label{fig:realworldresult}
\end{figure*}

\subsection{Simulated Experimental Result of Multitask Learning}
In each experiment, the agent is trained for around 10-20 minutes on a single computer (AMD Ryzen Threadripper 3960X, 24x 3.8GHz, NVIDIA GeForce RTX 2080 Ti, 11GB). The experiments conducted in this section do not apply domain randomization and RMA. All agent are implemented with Pytorch and tuned with Bayesian optimization.

\subsubsection{Multitask vs. Single-Task Learning} 
We first compare our Multitask SAC (MTSAC) trained on multiple tasks with a standard single-task SAC baseline. Specifically, we train the MTSAC agent on both attitude-tracking and stabilization tasksets (in total 57 tasks) simultaneously, while the single-task SAC and graph SAC agent are trained only on a single attitude tracking task. As shown in Fig.~\ref{fig:exp1singletask}, the multitask agent, when evaluated on the single task of angle tracking, converges more quickly and achieves more stable performance. This result highlights that learning additional tasks can improve exploration and robustness, thus accelerating training for attitude tracking task. Note-worthily, the single task Graph SAC demonstrate architecture advantage against plain architecture in single task learning. 

\subsubsection{Comparison with Baseline Multitask Architectures} 
\label{sec: Comparison with Baseline Multitask Architectures}
We next compare our GCN policy to other base models, including plain feed-forward, parallel, and compositional networks (Fig.~\ref{fig:exp3architecture}). The number of parameters for 24 neurons two layers are: Parallel 601, Compositional 655, Graph 697. For the plain feed-forward with similar number of parameters, we have 16 neurons two layers which has a in total 662 parameters. The graph neural network is slightly larger as there are a fix number of graph parameters $\norm{\mathcal{A}}\times\norm{\mathcal{S}}=3*18=54$. And the larger network should give graph network a strong advantage as the number of graph parameters does not scale with neurons in hidden layers but only with number of states and actions. 
Experiments show that our GCN-based approach converges faster when the network size is reduced to fit stringent onboard hardware constraints. However, we have noticed that other architectures seem to achieve a higher final accuracy, suggesting that a complex representation is needed to further improve the model.

\begin{figure*}
	\centering
	\includegraphics[width=0.97\linewidth]{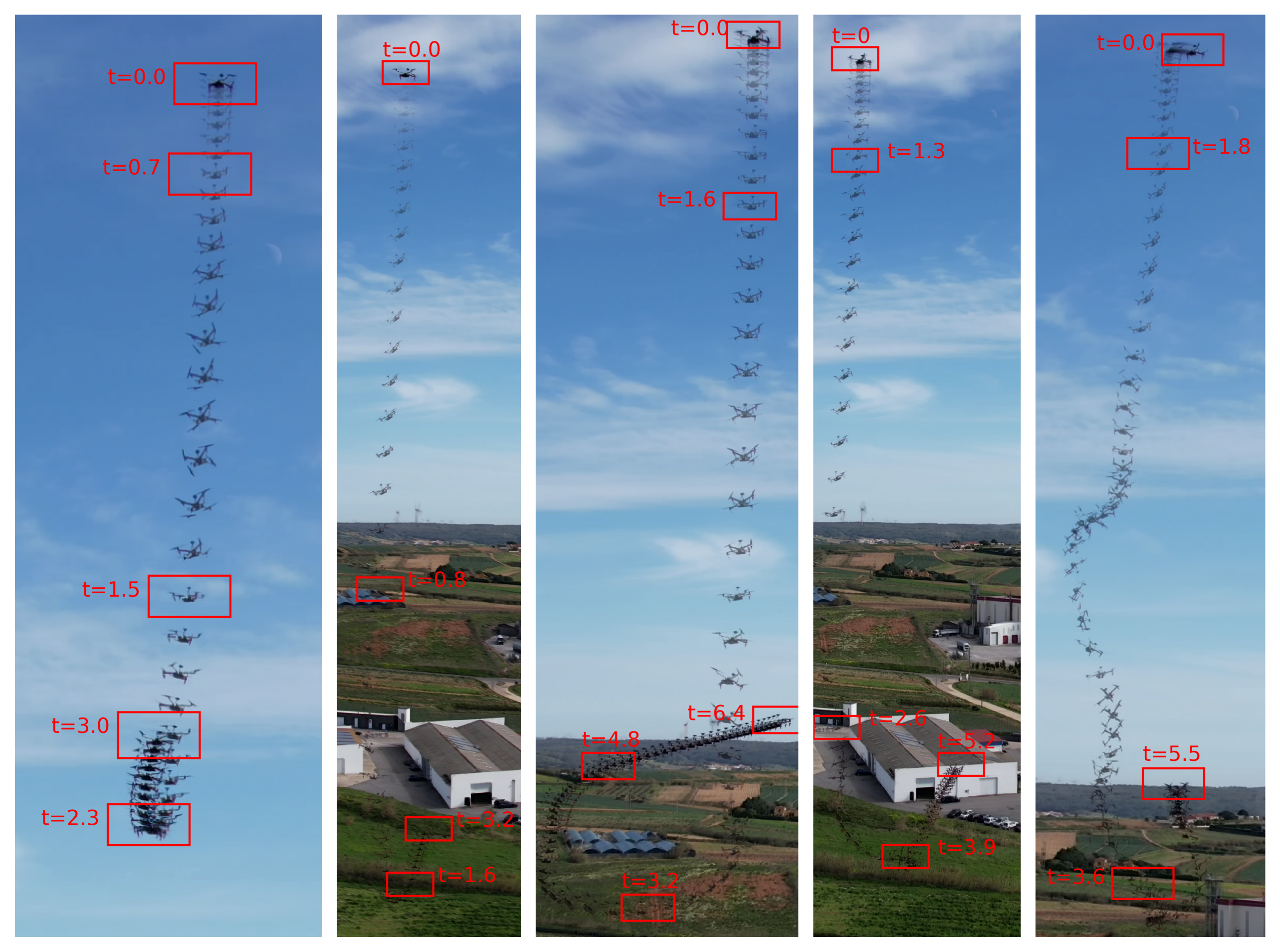}
	\caption{The footage of the real experiment. The rotor disengage around time zero for one to three seconds and the quadcopter starts to free-fall. Then the agent took over and re-engage the rotors. The agent is able to quickly adapt the situation and recover the pose even when the drone flips with high velocity and high angular velocity.}
	\label{fig:realworldfreefall}
\end{figure*}

\subsubsection{Effect of Task Diversity and Reset Conditions}
\label{sec: Effect of Task Diversity and Reset Conditions}
To evaluate the impact of task diversity on training efficiency, we trained agents under three distinct scenarios: (1) a small task set focused primarily on tracking-related tasks; (2) a medium task set aligned with the evaluation tasks for both tracking and stabilization; and (3) a large task set incorporating additional penalties for angular velocity and action smoothness (see Fig.~\ref{fig:exp4trainsets}). Agents trained on the medium and large task sets generally outperform those trained on the small set, indicating that a broader range of tasks promotes more effective exploration and enhances policy robustness.

Furthermore, we observe that after multitask training the agent modulates its behavior based on the provided task weights (see Fig.~\ref{fig:exp5multitaskproperty}). This behavior demonstrates that the policy has effectively learned multiple control modes—employing gentle maneuvers for smooth attitude tracking and more aggressive maneuvers for stabilization. However, despite yaw control being irrelevant to the stabilization task, the model continues to track the yaw angle target, suggesting that it cannot completely decouple the influence of irrelevant task components.

\subsubsection{Effect of Network Size on Graph MTSAC}
We finally evaluate the influence of network size on performance (Fig.~\ref{fig:exp6graphsacnetsize}). Although increasing the number of hidden units from 12 to 24 neurons per layer substantially improves sample efficiency, further expansion (to 36 neurons per layer) yields diminishing returns. This suggests that moderate-scale GCN architectures strike the best balance between computational cost and performance, which is crucial for onboard deployment.

\subsection{Simulated Experimental Results of Domain Transfer}

\subsubsection{Domain Adaptation with RMA} 
To evaluate robustness under varying dynamics, we integrate RMA (Sec.~\ref{sec:rma}) with domain randomization \cite{tobin2017domain}. Our network design, adapted from \cite{kumar2021rma}, reduces the adaptor size to meet onboard computational constraints. The encoder, with two layers of 128 neurons each, maps normalized environment factors (Table~\ref{tab: env latents}) to a two-dimensional latent vector, while the adaptor comprises a convolutional layer and a linear layer with 16 neurons (kernel size 8, padding 7, history length 50). For the two latent variables, $e_0$ connects to the roll and pitch nodes and $e_1$ to the yaw node; the final two columns in the trained graph (Fig.~\ref{fig:exp3graph}) confirm that the graph effectively captures the correlation between environment and action nodes.

Our results (Fig.~\ref{fig:exp7domainadaptation}) show that the RMA encoder significantly enhances adaptability compared to domain randomization alone. For real-world experiments, we train the adaptor to achieve a supervised loss below 1e-5 and run it at 4 Hz during deployment to reduce computational expense.

\subsection{Real-World Experiments using an F450 Quadcopter}
We evaluate our approach on an F450 quadcopter equipped with 930kV rotors and a Pixhawk-6X flight controller. The Ardupilot firmware provides high-level position and velocity control, while our trained low-level GCN policy operates on attitude commands. The policy is compiled in C++ and runs at approximately 400 Hz on the flight controller's onboard processor, namely STM32H753, whereas the RMA adapter operates at 4 Hz.

\subsubsection{Task 1: Attitude Tracking}
We first demonstrate that our multitask-trained agent can accurately follow attitude commands (see Fig.~\ref{fig:realworldtracking}). In this experiment, the high-level controller generates angle commands up to about 30 degrees. The RL agent tracks smoothly when the commanded angle is within 10 degrees, which lies within its training distribution. Notably, the agent exhibits near-perfect stability during both hovering and flight, maintaining a consistent attitude with minimal oscillations and rapidly damping any transient disturbances. Commands beyond this range tend to yield more conservative responses, suggesting that future training should incorporate a broader range of target angles.

\subsubsection{Task 2: Stabilization}
Next, we evaluate the stabilization capability of our approach. In these experiments, the rotors are disabled mid-flight to allow the vehicle to free-fall for one to three seconds before re-engaging the motors under the control of our GCN policy. The drone quickly stabilizes from high angular velocities, successfully recovering from extreme poses in all 13 trials (see Fig.~\ref{fig:realworldfreefallrollpitch} and Fig.~\ref{fig:realworldfreefall}).

	\section{Conclusions and Future Work}
\label{sec:5_discussion}

In this work, we presented a multitask deep reinforcement learning framework for quadcopter control that enhances exploration, accelerates convergence, and improves sample efficiency. By leveraging multitask learning and a GCN-based architecture to inject prior domain knowledge, our approach maintains a compact network size suitable for onboard deployment. Experimental results show that our framework converges faster than single-task methods, achieves stable training, and runs at 400 Hz on a Pixhawk flight controller without extra computing power, while also demonstrating robust stabilization under extreme conditions such as free-fall.

Nonetheless, the current model struggles to fully decouple irrelevant task components, and its compact design—mandated by stringent computational constraints—may limit the capacity to derive more complex representations, leading to lower final accuracy compared to alternative architectures. Future work will explore alternative graph construction methods, larger network architectures, and model compression strategies to better isolate task-specific behaviors, improve performance, and extend the framework to higher-level tasks such as velocity and position control.

	\printbibliography
\end{document}